\documentclass[runningheads]{llncs}
\usepackage[T1]{fontenc}
\usepackage{color, colortbl}
\definecolor{Gray}{gray}{0.9}
\definecolor{LightCyan}{rgb}{0.88,1,1}
\usepackage{graphicx}
\usepackage{hyperref}
\usepackage[dvipsnames]{xcolor}
\usepackage{amsmath}
\usepackage{amssymb}
\usepackage{cite}
\usepackage{arydshln}
\usepackage{multirow}
\begin{document}
\title{Leveraging Prompt-Tuning for Bengali Grammatical Error Explanation Using Large Language Models}
\titlerunning{Leveraging Prompt-Tuning for BGEE Using LLMs}
%
%\titlerunning{Abbreviated paper title}
% If the paper title is too long for the running head, you can set
% an abbreviated paper title here
%
\author{Subhankar Maity\orcidID{0009-0001-1358-9534} \and
Aniket Deroy\orcidID{0000-0001-7190-5040} }
%\and
%Sudeshna Sarkar\orcidID{0000-0003-3439-4282}}
%
\authorrunning{S. Maity \& A. Deroy}
% First names are abbreviated in the running head.
% If there are more than two authors, 'et al.' is used.
%
\institute{IIT Kharagpur, Kharagpur, India\\  
\email{\{subhankar.ai,roydanik18\}@kgpian.iitkgp.ac.in}}
%\email{sudeshna@cse.iitkgp.ac.in}}
%\email{subhankar.ai@kgpian.iitkgp.ac.in}, %\email{roydanik18@kgpian.iitkgp.ac.in}, 
%\email{sudeshna@cse.iitkgp.ac.in}}
%\url{http://www.springer.com/gp/computer-science/lncs} \and
%ABC Institute, Rupert-Karls-University Heidelberg, Heidelberg, Germany\\
%\email{\{abc,lncs\}@uni-heidelberg.de}}
%
\maketitle              % typeset the header of the contribution
\begin{abstract}
We propose a novel three-step prompt-tuning method for Bengali Grammatical Error Explanation (BGEE) using state-of-the-art large language models (LLMs) such as GPT-4, GPT-3.5 Turbo, and Llama-2-70b. Our approach involves identifying and categorizing grammatical errors in Bengali sentences, generating corrected versions of the sentences, and providing natural language explanations for each identified error. We evaluate the performance of our BGEE system using both automated evaluation metrics and human evaluation conducted by experienced Bengali language experts. Our proposed prompt-tuning approach shows that GPT-4, the best performing LLM, surpasses the baseline model in automated evaluation metrics, with a 5.26\% improvement in F1 score and a 6.95\% improvement in exact match. Furthermore, compared to the previous baseline, GPT-4 demonstrates a decrease of 25.51\% in \textit{wrong error type} and a decrease of 26.27\% in \textit{wrong error explanation}. However, the results still lag behind the human baseline.

\keywords{Bengali Grammatical Error Explanation (BGEE) \and Language Learning \and Prompt Tuning \and Large Language Models (LLMs)}
\end{abstract}

\section{Introduction and Background}

Generative AI has shown remarkable potential in transforming various natural language processing tasks \cite{r4}. Bengali, the seventh most spoken language globally \cite{r5}, possesses a rich linguistic structure, yet existing research in Bengali Grammatical Error Correction (BGEC) \cite{r6, r7, r9, r10, r11} struggles to provide comprehensive, learner-friendly explanations alongside error corrections. This limitation reduces its effectiveness in educational contexts. Developing robust error correction and explanation techniques for Bengali offers tremendous opportunities to advance education and language learning \cite{r2}. By leveraging LLMs, these systems can deliver interactive feedback, fostering improved educational tools and personalized learning experiences \cite{r2}. Such innovations align with the goals of the Artificial Intelligence in Education (AIED) community, emphasizing AI's role in enabling meaningful learning experiences. Prompt tuning \cite{r23} has shown remarkable performance in several tasks \cite{r26}, from multi-label text classification \cite{r24, r25} to various computer vision tasks \cite{r22,r27}. However, no work has explored the capability of prompt-tuning for the grammatical error explanation (GEE) task. In this paper, we address the Bengali grammatical error explanation (BGEE) task by introducing a novel three-step prompt-tuning approach for LLMs such as GPT-4 \cite{r29}, GPT-3.5 Turbo, and Llama-2-70b \cite{r30} to categorize grammatical error types, generate corrected sentence, and natural language explanations for grammatical errors in Bengali sentences. We leverage an existing dataset \cite{r2} comprising erroneous sentences, their corresponding correct sentences, and error types to prompt-tune these LLMs. Our work explores the potential of LLMs to advance the state-of-the-art in BGEE task, contributing to better information retrieval and educational support for Bengali-speaking communities. As contributions, (\textit{i}) We propose a novel three-step prompt-tuning approach using state-of-the-art LLMs such as GPT-4, GPT-3.5 Turbo, and Llama-2-70b for improving grammatical error correction and explanation for bengali language; (\textit{ii}) We evaluated the performance of the BGEE system using automated evaluation, as well as human evaluation by appointing experienced Bengali language experts.\\

\noindent \textbf{State-of-the-Art.}
Although there is increasing attention towards GEC in high-resource languages such as English \cite{r12, r13, r14}, Chinese \cite{r15}, German \cite{r16}, Russian \cite{r17}, Spanish \cite{r18}, etc., there is a distinct lack of research focused on GEC for low-resource languages such as Bengali. While there has been prior GEC research for Bengali \cite{r6, r7, r9, r10, r11}, the areas of feedback and explanation generation remain unexplored in this context. A significant contribution by \cite{r2} in the field of GEE for the Bengali language involves the use of one-shot prompted LLMs. However, their work is still in the preliminary stages and does not perform well for all types of Bengali grammatical errors. Our GEE task aims to fill this gap by focusing on the three step prompt-tuning of LLMs, followed by both automatic and human evaluation for the Bengali language.

\section{Task Definition} \label{task}
The BGEE task involves prompt-tuning LLMs to generate natural language explanations for grammatical errors in Bengali sentences. Specifically, given an erroneous sentence, the model must: (\textit{1}) Identify and categorize the grammatical errors; (\textit{2}) Generate a corrected version of the sentence; (\textit{3}) Provide a natural language explanation for each identified error.

\section{Dataset} \label{data}
The BGEE Dataset \cite{r2} consists of (\textit{i}) Erroneous Sentence: \(S_{\text{err}} = \{w_1, w_2, w_3, ..,\\ w_n\}\); (\textit{ii}) Correct Sentence: \(S_{\text{corr}} = \{w'_1, w'_2, w'_3, .., w'_m\}\), the grammatically correct version of the \(S_{\text{err}}\); (\textit{iii}) \(E_{\text{types}}\) the categorization of grammatical errors present in \(S_{\text{err}}\). The dataset is structured as $\{S_{\text{err}}, S_{\text{corr}}, E_{\text{types}}\}_{i=1}^N$, where $N$ is the number of triples. The dataset contains several Bengali error types \cite{r2} such as spelling, orthography, case-marker, subject-verb agreement, auxiliary verbs, pronouns, Guruchondali dosh, punctuation, verb tense, word order, etc. \cite{r2} categorizes these error types into three cognitive levels: \textit{single-word level errors}, \textit{inter-word level errors}, and \textit{discourse-level errors}. As the dataset proposed by \cite{r2} does not contain explicit explanations for the error types, we appointed five Bengali language experts through \href{https://www.surgehq.ai/faq}{Surge AI} to generate explanations for each triple in the dataset. Each explanation for a triple is denoted as \(S_{\text{explain}}\). The entire task was divided among the five experts. After annotation, the whole dataset is structured as $\{S_{\text{err}}, S_{\text{corr}}, E_{\text{types}}, S_{\text{explain}}\}_{i=1}^N$, where $N$ is the number of quadruples.

\section{Methodology} \label{meth}
\begin{figure}[h!]
  \centering
  \includegraphics[width=0.70\linewidth]{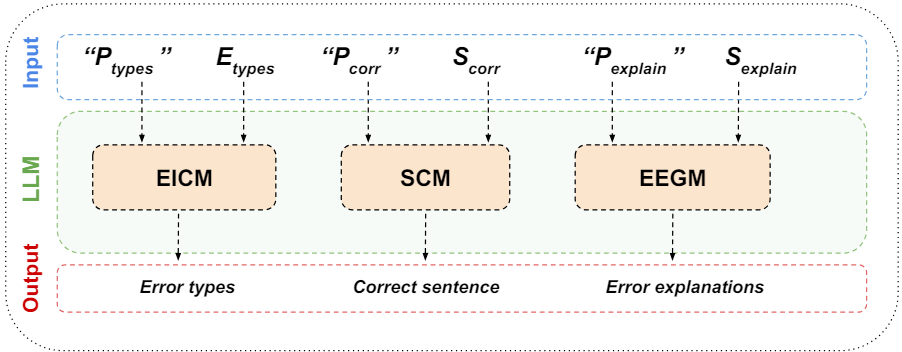}
  \caption{Overview of the proposed LLM prompt-tuning strategy. \textbf{LLM} denotes Large Language Model, \textbf{EICM} denotes the Error Identification and Categorization Module, \textbf{SCM} denotes the Sentence Correction Module, and \textbf{EEGM} denotes the Error Explanation Generation Module. The prompt fed to the LLM is denoted by “ ”. Definitions of the input notations (e.g., “\(P_{\text{types}}\)”, \(E_{\text{types}}\), etc.) are mentioned in Section \ref{meth}. } \label{fig_1}
  %\Description{A woman and a girl in white dresses sit in an open car.}
\end{figure}

%\subsection{Prompt-tuning Process}
The proposed prompt-tuning (PT) process, as shown in Fig. \ref{fig_1}, involves three primary components:

\noindent \textbf{1. Error Identification and Categorization Module (EICM)}: This module is responsible for detecting grammatical errors in the input sentences and classifying them into predefined categories. The input to this module consists of a prompt (denoted as “\(P_{\text{types}}\)”) designed to elicit error types and the corresponding gold standard error types (denoted as '\(E_{\text{types}}\)'). The prompt, \(P_{\text{types}}\), is structured as follows: 

“\textit{Provide the error types for the following erroneous Bengali sentence}. 

\{\textit{Erroneous sentence}\} 

\textit{Error types}:”

\noindent \textbf{2. Sentence Correction Module (SCM)}: This module generates the corrected version of the input sentence. The inputs to this module are the prompt for generating a grammatically correct sentence, “\(P_{\text{corr}}\)”, and the gold standard correct sentence, '\(S_{\text{corr}}\)'. The prompt (denoted as “\(P_{\text{corr}}\)”) used in this module is as follows: 

“\textit{Provide the grammatically correct sentence for the following erroneous Bengali sentence.} 

\{\textit{Erroneous sentence}\}

\textit{Correct sentence}:”

\noindent \textbf{3. Error Explanation Generation Module (EEGM)}: This module generates natural language explanations for the identified error types. The inputs to this module include the prompt for generating explanations, “\(P_{\text{explain}}\)”, and the gold standard explanations generated by Bengali language experts (as discussed in Section \ref{data}), '\(S_{\text{explain}}\)'. The prompt (denoted as “\(P_{\text{explain}}\)”) for this module is as follows: 

“\textit{Provide concise explanations for the types of grammatical errors in the erroneous Bengali sentence.}

\{\textit{Erroneous sentence, Correct sentence, Error types}\} 

\textit{Error explanations:}”

\noindent In addition, we conducted a comparison between the proficiency of prompt-tuned LLMs and that of four Bengali language experts (i.e., human baseline) recruited through \href{https://www.upwork.com/}{UpWork}. The set of erroneous sentences was partitioned into four portions for evaluation by each expert assigned to their respective portion. They are asked to perform the same three tasks (See Section \ref{task}) as the LLMs.

\section{Experiments}

We prompt-tune three LLMs for the BGEE task: GPT-4, GPT-3.5 Turbo (abbreviated as GPT-3.5), and Llama-2-70b (abbreviated as Llama-2). Following \cite{r19}, we use the number of epochs = 30 and keep the default values of other hyperparameters such as batch size, learning rate, etc.. We split the dataset (See Section \ref{data}) into 70\% for prompt-tuning the LLMs and 30\% for testing.

\noindent \textbf{Evaluation.} 
Following \cite{r1, r2, r3}, the automatic evaluation procedure involves examining performance at both the token level, which includes metrics such as precision, recall, $F_{1}$ score, and $F_{0.5}$ score and at the sentence level (i.e., exact match). The exact match (EM) evaluates the consistency between the predicted and reference sentences. Following \cite{r2, r20}, we also enlisted the expertise of another three experienced Bengali language instructors, recruited through \href{https://www.upwork.com/}{UpWork}, to evaluate the explanations (i.e., human evaluation). Each erroneous sentence, corrected sentence, and explanations generated by each LLM and a human expert were presented to one of the three instructors. They were tasked with identifying two types of errors in the explanations: \textit{wrong error type} \cite{r2, r20} (an error type not found in the erroneous sentence according to the gold standard error type) and \textit{wrong error explanation} \cite{r2, r20} (an error explanation not associated with the specific error type provided by human experts).

\noindent \textbf{Baselines.}
(\textit{i}) We compare the performance of prompt-tuned LLMs with a baseline model \cite{r2} that utilizes the one-shot prompting method across several LLMs for the BGEE task. Among the LLMs explored in \cite{r2} under the one-shot setting, we specifically compare with GPT-4 Turbo, GPT-3.5 Turbo, and Llama-2-70b, as these were identified as the best-performing LLMs according to \cite{r2}. (\textit{ii}) We also compare the performance of prompt-tuned LLMs with a human baseline (See Section \ref{meth}).

\section{Results}

\begin{figure}[h!]
  \centering
  \includegraphics[width=1\linewidth]{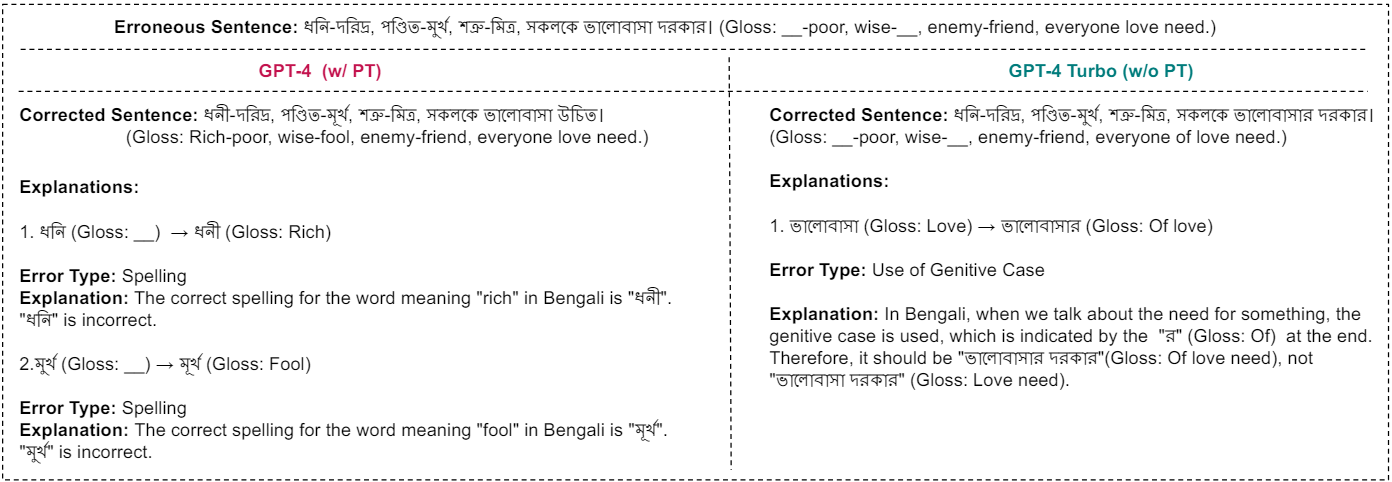}
  \caption{Example of an erroneous Bengali sentence (containing a spelling error) with \textcolor{purple}{GPT-4 (w/ PT)}'s GEE output and \textcolor{teal}{GPT-4 Turbo (w/o PT)}'s GEE output (baseline). PT denotes \textit{prompt-tuned}. “\textunderscore” in gloss denotes the spelling error in the Bengali word.} \label{fig_2}
  %\Description{A woman and a girl in white dresses sit in an open car.}
\end{figure}

Table \ref{tab_r} shows that prompt-tuning significantly enhances the performance of various LLMs, including GPT-4, GPT-3.5 Turbo, and Llama-2-70b, over baseline LLMs such as GPT-4 Turbo, GPT-3.5 Turbo, and Llama-2-70b, which employ one-shot prompts for the BGEE task. This improvement is observed across all error levels discussed in \cite{r2} and automated evaluation metrics, such as Precision, Recall, F\textsubscript{1}, F\textsubscript{0.5}, and exact match. Although the human baseline outperforms prompt-tuned LLMs due to their pre-training on smaller datasets from low-resource languages such as Bengali\footnote{\href{https://github.com/openai/gpt-3/blob/master/dataset_statistics/languages_by_word_count.csv}{Link to a .csv file comprising training data statistics for GPT-3} }. Following \cite{r21} we also calculate Pearson's $r$ between human experts and the best-performing LLM (i.e., GPT-4). Table \ref{tab_pear} indicates that F\textsubscript{0.5} achieves the highest Pearson's $r$ value. Table \ref{human_llm} further demonstrates that our proposed prompt-tuning approach significantly enhances human evaluation results compared to the baseline. As shown in Fig. \ref{fig_2}, GPT-4 (w/ prompt tuning) accurately detects spelling errors in the erroneous Bengali sentence and provides precise explanations. In contrast, GPT-4 Turbo (w/o prompt tuning) fails to identify spelling errors and incorrectly labels them as "Use of Genitive case", resulting in inaccurate explanations. Notably, prompt-tuning improves LLMs' ability to identify various Bengali grammatical error types, such as word order, spelling, case marker errors, and Guruchondali dosh, surpassing the capabilities of the previous baseline \cite{r2}. This advancement is attributed to the enhanced error identification and explanation facilitated by prompt-tuned LLMs compared to the one-shot prompts utilized in \cite{r2}.

\begin{table}
\centering
\renewcommand{\arraystretch}{1.0}
\caption{\label{tab_r}
Performance comparison in predicting grammatically correct Bengali sentences for various error types and overall. 'Human' denotes the human baseline, \textcolor{purple}{Purple} color represents prompt-tuned (PT) LLMs, and \textcolor{teal}{Teal} color denotes one-shot LLMs (baseline \cite{r2}). 'EM' denotes \textit{exact match}, and 'PT' denotes \textit{prompt-tuned}. 
}
\scalebox{0.60}{
\begin{tabular}{|c|c|c|c|c|c|c|c|}
\hline
\textbf{Metric} & \textbf{Human} & \textbf{\textcolor{purple}{GPT-4 (PT)}} & \textbf{\textcolor{teal}{GPT-4 Turbo}} & \textbf{\textcolor{purple}{GPT-3.5 Turbo (PT)}} &  \textbf{\textcolor{teal}{GPT-3.5 Turbo}} & \textbf{\textcolor{purple}{Llama-2-70b (PT)}} &  \textbf{\textcolor{teal}{Llama-2-70b}}\\ \hline 
\multicolumn{8}{|c|}{\textcolor{blue}{Single-word level errors}} \\ \hline 
Precision & 95.84 & 82.45 & 74.47 & 75.26 & 69.90 & 74.62 & 71.84 
\\ 
Recall & 91.77 & 74.27 & 72.81 & 70.47 & 66.81 & 72.35 & 68.90 
 \\ 
F\textsubscript{1} & 93.57 & 77.38 & 73.39 & 70.52 & 67.35 & 72.47 & 69.32 
 \\ 
F\textsubscript{0.5} & 94.94 & 79.86 & 73.81 & 71.42 & 68.79  & 74.70 & 70.61 
\\
EM & 74.44 & 52.53 & 48.69 & 43.88 & 39.62 & 49.61 & 45.30  \\  \hline 
\multicolumn{8}{|c|}{\textcolor{blue}{Inter-word level errors}} \\ \hline 

Precision & 91.44 & 72.31 & 68.84 &  68.53 & 62.91 & 65.70 & 63.72 
\\ 
Recall & 88.67 & 69.90 & 65.60 & 64.21 & 60.74 & 63.44 & 60.73 
 \\ 
F\textsubscript{1} & 89.20 & 71.25 & 66.39 & 68.33 & 61.99 & 63.58  & 61.49 
 \\ 
F\textsubscript{0.5} & 89.73 & 72.21 & 67.82 & 69.36 & 62.35  & 66.41 & 62.28   \\ 
EM & 69.21 & 50.11 & 46.70 & 46.72 & 43.91 & 47.88 & 45.80   \\  \hline 
\multicolumn{8}{|c|}{\textcolor{blue}{Discourse level errors}} \\ \hline 
Precision & 94.26 & 74.51 & 70.57 & 74.99 & 67.88 & 68.92 & 65.84 
\\ 
Recall & 89.21 & 70.22 & 67.75 & 68.15 & 65.81 & 65.90 & 62.83 
 \\ 
F\textsubscript{1} & 90.78 & 73.26 & 69.42 & 70.15 & 66.32 & 66.78 & 63.75
 \\ 
F\textsubscript{0.5} & 91.22 & 73.44 & 70.47 & 71.25 & 66.74  & 68.23 & 64.11
\\
EM & 71.49 & 54.31 & 50.71 & 49.94 & 46.83 & 48.26 & 45.92  \\  \hline 
\multicolumn{8}{|c|}{\textcolor{blue}{Overall}} \\ \hline
Precision & 93.48 & 76.88 & 71.11 & 73.21 & 66.79 & 71.25 & 66.85 
\\ 
Recall & 89.22 & 71.24 & 68.48 & 67.88 & 64.39 & 67.26 & 63.87 
\\ 
F\textsubscript{1} & 90.76 & 73.20 & 69.54 & 70.07 & 64.94 & 67.35 & 64.59 
 \\ 
F\textsubscript{0.5} & 92.12 & 74.22 & 70.54 & 71.22 & 65.85 & 70.11 & 65.36 \\ 
EM & 71.67 & 52.45 & 49.04 & 46.21 & 43.89 & 48.25 & 45.69  \\  \hline 
\end{tabular}
}
\end{table}

\begin{table}
\centering
\renewcommand{\arraystretch}{1.0}
\caption{\label{tab_pear}
Pearson's $r$ between the top-performing LLM (i.e., GPT-4) and human experts across various automated evaluation metrics. 'EM' denotes \textit{exact match}.
}
\scalebox{0.65}{
\begin{tabular}{|c|c|c|c|c|}
\hline
\textbf{Precision} & \textbf{Recall} & \textbf{F\textsubscript{1}} & \textbf{F\textsubscript{0.5}} & \textbf{EM}\\
\hline 
0.582 & 0.544 & 0.561 & \textbf{0.590} & 0.529 \\
\hline
\end{tabular}
}

\end{table}

\begin{table} 
\centering
\renewcommand{\arraystretch}{1.0}
\caption{\label{human_llm} 
Human evaluation results of various LLMs for BGEE. \textcolor{purple}{Purple} color represents prompt-tuned (PT) LLMs, and \textcolor{teal}{Teal} color denotes one-shot LLMs (baseline \cite{r2}). 'WET' represents the \textit{wrong error type}, 'WEE' represents the \textit{wrong error explanation}, and 'PT' represents \textit{prompt-tuned}.  
}
\scalebox{0.60}{
\begin{tabular}{|c|c|c|c|c|c|c|}
\hline
\textbf{Metric} & \textbf{\textcolor{purple}{GPT-4 (PT)}} & \textbf{\textcolor{teal}{GPT-4 Turbo}} & \textbf{\textcolor{purple}{GPT-3.5 Turbo (PT)}}& \textbf{\textcolor{teal}{GPT-3.5 Turbo}} & \textbf{\textcolor{purple}{Llama-2-70b (PT)}} & \textbf{\textcolor{teal}{Llama-2-70b}} \\ 
\hline 
%\multicolumn{11}{c}{\textit{\textbf{Single-word level error}}} \\ \hline 
WET (\%) & 20.35 & 27.32 & 25.44 & 30.37 & 27.41 & 33.19 
\\ 
WEE (\%) & 26.46 & 35.89 & 30.40 & 38.82 & 32.11 & 39.04 
\\ 
\hline 
\end{tabular}
}

\end{table}

\section{Conclusion and Future Work}

In conclusion, our proposed prompt-tuning approach significantly enhances the performance of LLMs in the BGEE task. Our rigorous evaluation demonstrates notable improvements over baseline LLMs across diverse error types and evaluation metrics. Importantly, our approach, led by GPT-4, excels in both automated and human evaluations, demonstrating improvements in error identification, providing grammatically correct sentence and explanation generation. Prompt-tuned GPT-4 outperforms the baseline model in automated evaluation metrics with a 5.26\% improvement in F1 score and a 6.95\% improvement in exact match. Additionally, compared to the previous baseline, it demonstrates a 25.51\% reduction in wrong error type and a 26.27\% reduction in wrong error explanation. This highlights the efficacy of prompt-tuning in improving LLM performance, particularly in identifying various Bengali grammatical error types such as word order, spelling, case marker errors, and Guruchondali dosh, surpassing the previous baseline. However, our findings also underscore the persistent gap between LLMs and human baseline in the BGEE task, necessitating further research to refine LLM applications for GEE in Bengali and beyond.

% ---- Bibliography ----
%
% BibTeX users should specify bibliography style 'splncs04'.
% References will then be sorted and formatted in the correct style.
%

\bibliographystyle{splncs04}
\bibliography{ref.bib}

\end{document}